\documentclass{frey}

\usepackage[english]{babel}
\usepackage{xcolor}
\usepackage{anyfontsize}
\definecolor{darkred}{RGB}{195, 32, 20}

\title{Advances, challenges, and opportunities for legged robots}

\author[1,2,3]{Jonas Frey$^{\mathbf{1,2,3}}$, Mat{\'i}as Mattamala$\mathbf{^{4}}$, Hae-Won Park$^\mathbf{{5}}$, Mayank Mittal$^{\mathbf{1,6}}$, Georg Martius$^{\mathbf{7,8}}$, Maike Osborne$^{\mathbf{9}}$, Robert~Sparrow$^{\mathbf{10,11}}$, Marco Hutter$^{\mathbf{1,12}}$}

\email{jonfrey@stanford.edu; matias.mattamala@ed.ac.uk; mahutter@ethz.ch}

\address{%
	Robotic Systems Lab\\
	ETH Zurich\\
	Zurich\\
	Switzerland
}

\contact{%
	Dr. Jonas Frey\\
	E-mail: jonfrey@stanford.edu\\
	Dr. Mat{\'i}as Mattamala\\
	E-mail: matias.mattamala@ed.ac.uk\\
	Prof. Marco Hutter\\
	E-mail: mahutter@ethz.ch
}

\program{}

\keyword{LEGGED ROBOTS}

\begin{document}

\abstract{Humanoid and quadrupedal robots have the potential to revolutionize the way we work, interact, and coexist with intelligent machines. To understand their effects on society and how they can enable scientific discovery, we assess the current capabilities of these systems along hardware, locomotion, autonomy, data, and applications. We identify recent advances and key open challenges that must be overcome to enable widespread adoption and new use cases for legged robots. Last, we provide an outlook on the future of legged robots, exploring their ethical considerations, economic potential, policy implications, and broader societal effects.
}
\maketitle

\section{Introduction}
\label{sec:introduction}
Legged robots, once discussed mainly in the realm of science fiction, have undergone major technological advances in recent years, enabling them to deliver parcels directly to front doors \cite{rivr2025, agility2025digit} or support proof-of-concept search-and-rescue demonstrations \cite{tranzatto2022cerberus}. What was once regarded as purely speculative is now becoming reality. We are witnessing a surge in financial investment and academic research and the rapidly growing robustness and versatility of contemporary humanoid and quadrupedal robots, opening up broad socioeconomic opportunities.

Technically, legged robots are inspired by the biomechanics of animals and use legs to locomote over challenging terrains. Compared with wheels, legs allow stepping over obstacles and climbing up staircases while offering active suspension and dynamic stabilization using only a small footprint \cite{raibert1985legged}. Whereas aerial drones can suffer from short operation times and the challenges of manipulating objects in the air, legged systems can operate for extended periods, carry heavy payloads, and even perform precise mobile manipulation. Furthermore, their unique morphology renders them ideal for operation in human-centric environments.

The development of legged robots has evolved considerably since their origins in the past century \cite{raibert1985legged, todd1985walking, devin2023optimizing}. Machines like the hydraulically powered General Electric Walking Truck were teleoperated platforms aimed to carry heavy loads over challenging terrain. Developments in the late 1970s, such as the Ohio State University hexapod, used them as developmental platforms, laying the groundwork for digital control. In Japan, Ichiro Kato was a pivotal figure, developing early humanoid robots like the WABOT-1 (WAseda roBOT). This work focused on creating full-scale anthropomorphic machines. Another shift in the field came from the work of Marc Raibert \cite{raibert1985legged}, moving the focus from statically stable legged robots toward dynamic stability. This progress led to Honda's series of humanoid robots, with the P2 becoming the first self-contained bipedal robot to use an onboard computer for walking, a predecessor to Honda's ASIMO (Advanced Step in Innovative Mobility) \cite{sakagami2002asimo}, the most well-known humanoid of its time. Recently, robots like Boston Dynamics' LS3 (Legged Squad Support System) and BigDog \cite{raibert2008bigdog} have demonstrated the consolidation of dynamically stable quadruped machines. Still, in the past decade, we have seen an explosive increase in the capabilities, diversity, and access to legged platforms. How did we get here? What is missing? What should we do next? How will the world change?

Here, we review the advances in legged hardware and machine learning techniques that have made it possible, leading to a surge in both academic research and industry activity. Hardware advancements, such as reliable and low-cost actuation, miniaturization of sensors, improved onboard computation, and open-source designs \cite{katz2018low, Grimminger}, have led to increased availability of legged robots. This, combined with the fast-paced progress in machine learning, particularly reinforcement learning (RL) \cite{hwangbo2019learning, tan2018simtoreal}, and an evolving open-source software ecosystem \cite{todorov2012mujoco, makoviychuk2021isaacgym ,rudin2022learning , mittal2025isaaclab, zakka2025mujocoplayground}, have been transformative for the field. These developments have unlocked advanced control and locomotion skills, allowing robots to evolve from bulky machines constrained to laboratory or industrial settings to agile systems capable of walking outdoors autonomously, with an unprecedented level of environment understanding.

As legged robots are increasingly adopted in different sectors, their effects extend beyond pure research to the economy and society. The value of this Review lies not merely in summarizing technical advances, highlighting open challenges, and discussing current real-world applications but also in reflecting on these developments at a time when their societal, economical, and ethical implications are increasingly critical. Grounded in technical progress and its anticipated socioeconomic repercussions, we provide actionable recommendations for developers, policy-makers, legislators, and other decision-makers.

\begin{figure*}
	\centering
	\includegraphics[width=1.0\textwidth]{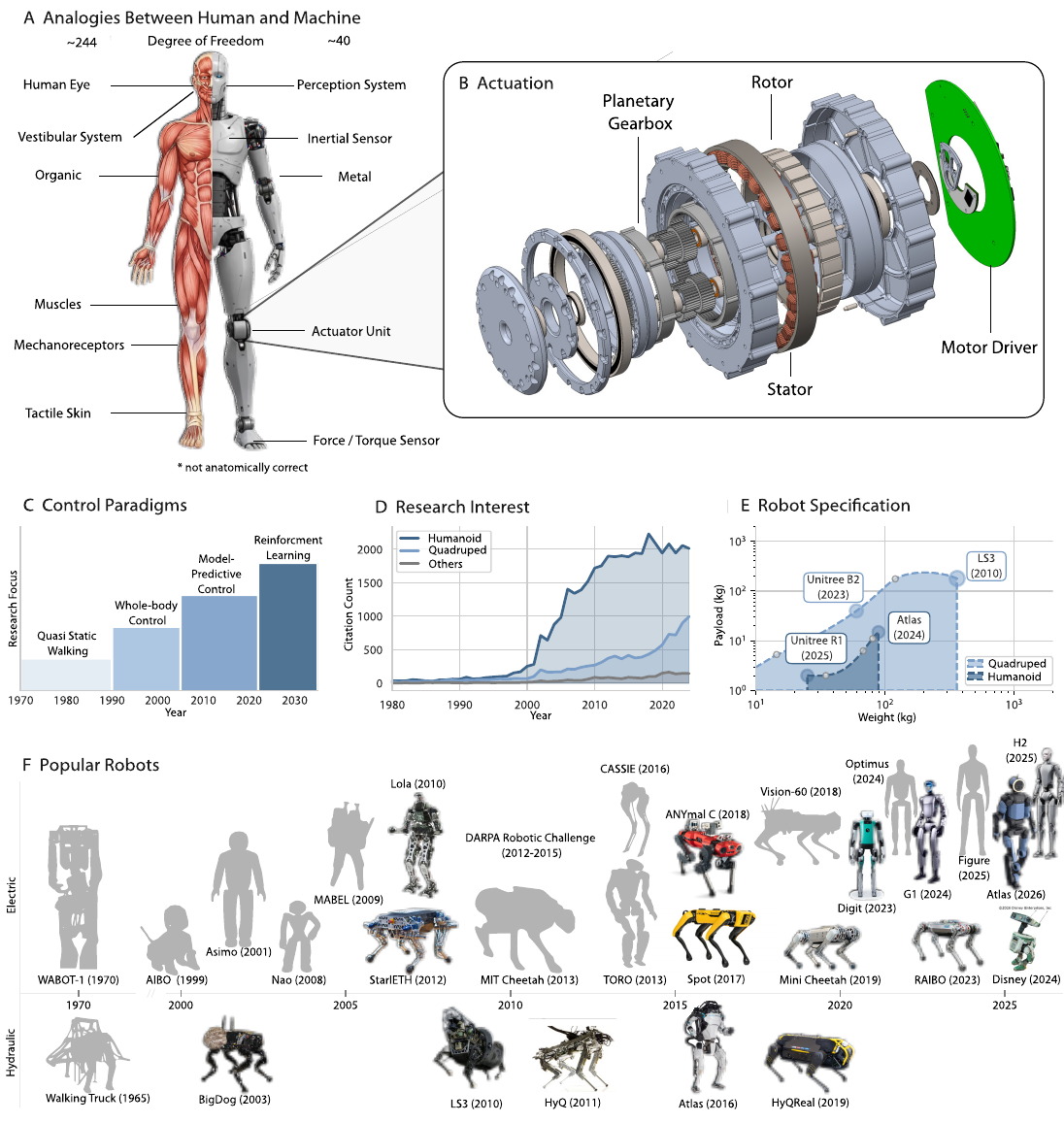} 
	\caption{\footnotesize\textbf{Hardware overview.}
	 (\textbf{A}) A comparison of the functional analogies between human body parts (for example, eye and muscles) and robot components (perception and actuation). (\textbf{B}) The internal structure of a robotic actuation unit \cite{jonhun2025design}. (\textbf{C}) Historic development of the control paradigm toward RL. (\textbf{D}) The growing research interest in humanoid and quadrupedal robots \cite{openalex_humanoids, openalex_quadrupeds, openalex_others}. (\textbf{E}) Payload capacity versus weight of popular quadrupeds and humanoids as of 2025. (\textbf{F}) Selected milestones in the development of legged robots in terms of applications, capabilities, or technologies.}
	\label{fig:hardware}
\end{figure*}

\section*{Hardware}
\label{sec:hardware}
The hardware of legged robots consists of actuators, sensors, electrical components, and mechanical structures that define the robot's morphology. Quadrupedal and bipedal robots are the most common morphologies (Fig.~\ref{fig:hardware}). Quadrupeds transitioned from statically to dynamically stable locomotion, making them well suited for load carrying or manipulation, with payload capacities of quadrupedal robots ranging up to 180 kg (Fig.~\ref{fig:hardware}E). Conversely, humanoids offer a large workspace for manipulation, but their reliance on dynamic stability makes both manipulation and locomotion more challenging. Functionally, the robot's hardware operates through a continuous sense-act cycle: It processes heterogeneous sensor measurements to estimate the robot's proprioceptive state and external environments. This information is then used to compute actuator commands. Actuators convert these commands into joint torques, which accelerate the robot's multibody system and result in contact forces at the feet or hands. This closed interaction enables the robot to achieve fast, reactive control actions and high-level decisions to accomplish locomotion tasks.

A key requirement for actuators is the capability to generate sufficiently high torque to create desired motions while supporting the robot's body weight. When running, the vertical ground reaction force in quadrupedal robots can reach up to three times their body weight \cite{park2017highspeed}, requiring large joint torques. Furthermore, during instantaneous contact events with the terrain, actuators must rapidly respond to large impulsive impacts \cite{wensing2017proprioceptive, siciliano2016handbook}. These challenges require high-torque actuators with low mechanical impedance and the resulting high backdrivability---requirements that differ from those of industrial robot arms \cite{pratt1995series}.

Legged robots have typically used electric motors. Because electric motors were historically designed for high-speed output because of power efficiency and manufacturing considerations, older motors provide insufficient torque and therefore require large gear reductions \cite{hollerbach1992comparative, pratt1995stiffness}. However, high gear ratios inevitably introduce friction, large reflected rotor inertia, and increased mechanical impedance \cite{pratt1995stiffness}. As a result, the actuators become nonbackdrivable, and torque control at the joints is not feasible without additional sensors or mechanisms \cite{englsberger2014overview}. These limitations make the joints vulnerable to ground impacts and limit the robot's capacity for dynamic locomotion. To address these issues, series elastic actuators were proposed, in which a spring is connected after the gear train to compliantly interact with the environment \cite{pratt1995series}, and have since been incorporated into legged robot designs \cite{pratt2008designbipedal,hutter2013starleth}. However, the compliance reduces control bandwidth, because resonance lowers achievable closed-loop stiffness, impairs high-frequency torque tracking, and is difficult to model \cite{robinson2000series}. Alternatively, hydraulic actuators offer a high power-to-weight ratio compared with electric motors. Early legged robots from the MIT Leg Laboratory demonstrated highly dynamic motions with hydraulics \cite{raibert1985legged, raibert1990trotting}. Later platforms from Boston Dynamics, such as BigDog \cite{raibert2008bigdog} and the Atlas series \cite{bostondynamics2016atlas}, integrated compact pumps, valves, and fluid reservoirs directly into the robot. However, hydraulic systems are hindered by mechanical complexity, high production and maintenance costs, low efficiency, and practical issues such as noise and oil leakage \cite{semini2011design,hutter2016anymal}. This has led to a decreased interest in their development (Fig.~\ref{fig:hardware}F).

More recently, high-torque, backdrivable actuation with electric motors has been achieved through the use of custom motors, characterized by a topology with a large gap radius and a short axial length, paired with low gear-ratio transmissions \cite{seok2012actuator, wensing2017proprioceptive}, illustrated in Fig.~\ref{fig:hardware}B. This combination yields low mechanical impedance, which in turn enables high backdrivability. Because of high transparent gear transmission, joint torque can be closely approximated as linearly proportional to motor current, scaled by the motor torque constant and gear ratio \cite{seok2012actuator}. As a result, neither additional force or torque sensors nor complex control loops are required to achieve precise, high-bandwidth torque control. Furthermore, supported by open-source releases of actuator designs \cite{katz2018low}, there has been a boom in quadruped and humanoid design adopting this design philosophy (see Fig.~\ref{fig:hardware}F). Notable examples include Unitree's lineup of quadrupedal \cite{unitree2020a1} and humanoid platforms \cite{unitree2024g1}, as well as University of California, Los Angeles's ARTEMIS humanoid \cite{11203020}.

In parallel, the miniaturization of sensors has made it possible to integrate a wide range of sensors into legged robots, enabling comprehensive state estimation and environment awareness. Proprioceptive sensors, which measure internal body state variables, include joint encoders that measure joint angles and inertial measurement units that provide angular velocity and linear acceleration of the robot body \cite{siciliano2016handbook}. By fusing these signals, the robot can estimate its body orientation and motion \cite{bloesch2012state}. Exteroceptive sensors, including vision sensors, enable environmental perception. LIDAR (light detection and ranging) sensors provide dense three-dimensional (3D) point clouds, whereas RGB-D (red, green, blue, and depth channel) cameras combine visual and depth information. These sensors are widely used for terrain mapping and environment reconstruction \cite{frankhauser2018robot, bostondynamics2021atlas}, allowing the robot to anticipate terrain features, plan foot placements, and avoid obstacles \cite{miki2022learning, kumar2021rma}. Contact and force sensing provide information about the interaction between the feet and the ground: Simple binary sensors can detect touchdown and liftoff events \cite{park2011identification, hutter2013starleth}, whereas multiaxis force/torque sensors quantify ground reaction forces and moments \cite{jung2018development}. Such information can be used to improve slip detection and unexpected collisions \cite{park2013finite}. Ongoing research on tactile skin \cite{liu2022neuro} can further improve haptic feedback, allowing robots to perceive contacts along the body and improve their navigation capabilities. However, the limited durability and integration challenges of tactile skin have impeded its wider adoption.

The development of alternative actuators offers prospective avenues beyond electric motors. For example, biologically inspired muscles \cite{buchner2024electrohydraulic, hu2023bioinspired} can potentially offer advantages such as a smaller form factor, softness, and a higher force-to-weight ratio. However, these technologies are still developing and face issues with manufacturing, durability, and control, rendering them not yet viable in commercial systems. Similar challenges are faced by tendon-driven actuation, which has the potential to offer greater anatomical fidelity to human musculoskeletal structure \cite{doi:10.1126/scirobotics.aaq0899}. Another ongoing research area is transitioning from rigid designs to softer and more compliant morphologies, which offer improved safety and human interaction \cite{yasa2023soft}.

\begin{figure*}
	\centering
	\includegraphics[width=1.0\textwidth]{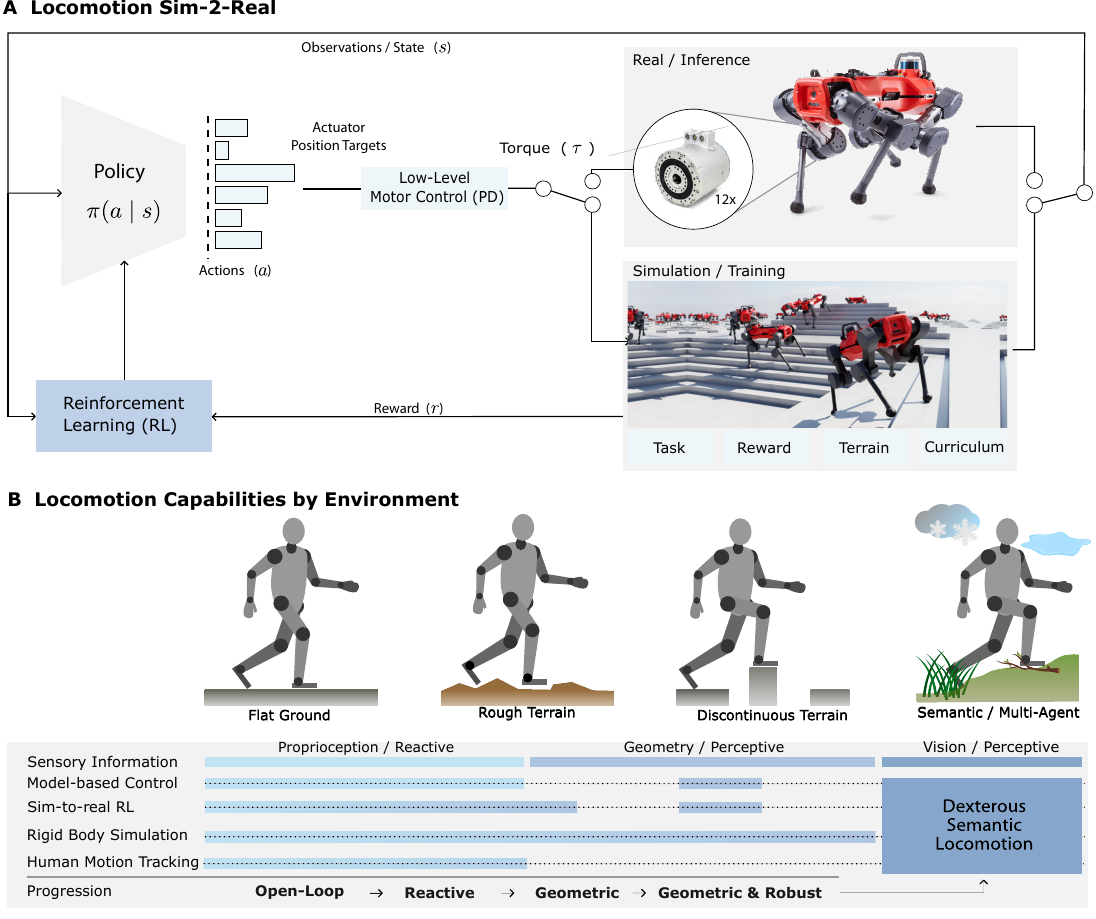}
	\caption{\footnotesize \textbf{Overview of legged locomotion.}
(\textbf{A}) The two main phases of the sim-to-real paradigm. During training, a policy is optimized via RL in a simulator to output actuator position targets. A motor controller typically converts these targets into the required torques, which are provided to the rigid body--physics simulator to compute the next state, observation, and reward. Through the use of domain randomization and system identification, the policy becomes robust enough to transfer to the real world without additional real-world training.
(\textbf{B})
Locomotion in complex environments requires increasingly sophisticated sensing, progressing from proprioceptive to geometric and lastly to visual information. Model-based control methods are highly effective when system dynamics and environmental interactions are well captured, yet their robustness can degrade in the presence of terrain-induced disturbances. RL policies trained in rigid-body simulators have demonstrated resilience in these challenging settings. Despite these advancements, a gap remains: Although current robots reliably navigate flat ground and increasingly traverse rough and discontinuous terrains, they lack semantic understanding of their environment and dexterity for careful and environment conditioned foot placement. Combining those two capabilities has the potential to account for social norms, multiagent interactions, and collaborative behaviors. We term this emerging research paradigm ``dexterous semantic locomotion.''
}
	\label{fig:locomotion}
\end{figure*}

\section*{Locomotion}
\label{sec:locomotion}
For a legged robot to efficiently use its hardware, algorithms must translate task specifications into motor commands, enabling effective interaction with the environment. These algorithms define the robot's motion, agility, and intelligence, ultimately determining its viable applications. Differences in sensor update rates, task latency requirements, and computational demands naturally lead to a modular control structure: fast lower-level modules (for example, actuator control at $\sim$200 to 1000 Hz, locomotion at $\sim$50 to 200 Hz) and slower higher-level modules ($<$30 Hz) that facilitate autonomy through scene understanding, planning, and navigation \cite{siegwart2011introduction}.

The locomotion controller typically generates actuator target positions that are passed to the actuator controllers to convert to motor currents or joint torques. It manages tasks such as foot placement, gait sequencing, interleg coordination, liftoff timing, and balancing to follow a specified velocity, position, or reference motion. The controller defines the motion of the robot by processing proprioceptive inputs---including joint positions, velocities, torques, and inertial data---alongside exteroceptive terrain perception.

Traditionally, classical methods decomposed locomotion into individual hand-engineered skills. Existing literature covers both the biomechanics \cite{holmes2006, navvad2018energy} and model-based control strategies \cite{reher2021dynamic} for legged locomotion. Early approaches focused on static stability, using the center of mass (COM) criterion, which requires the COM projection to remain strictly in the robot's support polygon. This restriction was later eased with the development of the zero-moment point criterion, followed by the adoption of reduced-order models such as the linear inverted pendulum and the spring-loaded inverted pendulum, enabling dynamic locomotion. Model predictive control (MPC) further enabled walking, jumping, and running behaviors by using simplified robot models \cite{reher2021dynamic}. However, the requirement for real-time control restricts model expressiveness, thereby limiting the controller's robustness under the stochasticity and uncertainty of real-world interactions. This limitation motivated approaches such as joint-level reflexes and other heuristics to increase robustness \cite{semini2015towards}. With the rise of RL-based controllers, MPC methods have motivated the development of hybrid systems, where MPC helps in long-horizon planning during training and deployment \cite{jenelten2024dtc}.

Policies trained with RL entirely in simulation have become a widely used paradigm for developing locomotion controllers \cite{ha2025learning} and fall recovery strategies \cite{huang2025standup}. They amortize and distill thousands of hours of simulated experience into a robust policy, predominantly using proximal policy optimization (PPO) during training \cite{schulman2017proximal}.

RL-based locomotion controllers have been successful because of crucial design decisions. The simulation environment reflects potential states that the robot will experience during deployment. The observation space is deliberately limited, often favoring geometric exteroception over RGB images. The policy predicts actuator position targets, converted to torques via impedance control rather than predicted directly, which eases exploration during training (Fig.~\ref{fig:locomotion}A).

However, the key enablers have been the techniques to mitigate the ``sim-to-real'' gap. System identification of the actuator models \cite{hwangbo2019learning} enabled the learned policies to transfer zero-shot to real hardware. Domain randomization \cite{tan2018simtoreal} yields policies that are resilient to variations, although at the cost of optimality, because the learned behavior is an average solution not tailored to the specific robot and environment \cite{lee2020learning}. Domain adaptive policy learning \cite{miki2022learning, kumar2021rma} addresses this shortcoming by continuously estimating system parameters online, allowing the robot to adapt to its environment and better cope with sensor noise and uncertainties in the real world.

To meet the stringent real-time demands and ease generalization and optimization, RL locomotion policies are implemented with low-capacity models, such as multilayer perceptrons \cite{hwangbo2019learning} or recurrent architectures \cite{miki2022learning} totaling below 10 million parameters. Although some studies have applied diffusion models \cite{huang2024diffuseloco} or transformer architectures \cite{radosavovic2024realworld}, these models are smaller than their vision and language counterparts, with millions instead of billions of parameters \cite{radford2021clip, lbmtri2025, kim2024openvla}.

RL often requires extensive reward shaping to produce locomotion behaviors. This can be alleviated by incorporating domain knowledge of symmetries \cite{abdolhosseini2019symmetry} or by using approaches including bioinspired central pattern generation \cite{bellegarda2022CPG}, constrained RL \cite{chanesane2024cat}, and automatic curriculum design \cite{wang2019poet}. Furthermore, the sample inefficiency of algorithms like PPO renders exploration difficult, leading to student-teacher methods that use privileged information \cite{chen2019learningcheating, lee2020learning, miki2022learning, zhuang2024humanoid}, yet the challenge of learning from limited data and real-world adaptation persists. Sample-efficient alternatives, such as off-policy \cite{haarnoja2019learning,wu2022daydreamer} or offline RL methods \cite{haarnoja2018sac, abdolmaleki2018mpo}, have not yet proven effective in this domain.

Defining the locomotion objective itself, such as velocity tracking, can be overly restrictive and prohibit learning of complex behaviors, such as jumping over gaps. To overcome this, promising research directions include optimizing for sparser position reward tasks \cite{rudin2022learning}, foothold tracking \cite{he2025attention}, and using imitation learning with human or animal reference data to develop diverse locomotion styles \cite{Peng2018deepmimic, Peng2021AMP, liao2025beyond, he2025asap} or agile behavior \cite{li2023wasabi, xie2025kungfubot}. Retargeting reference motions, whether sourced from video or motion capture, to a specific robot embodiment with varying limb lengths, mass distributions, and degrees of freedom is typically formulated as an optimization problem via inverse kinematics to align 3D joint positions and orientations. Compared with real-time retargeting, which is often limited by computational constraints, offline retargeting enables a higher degree of physical realism and motion fidelity \cite{ze2025twist}. The resulting retargeted motion is commonly tracked by an RL controller \cite{liao2025beyond, ze2025twist, he2025asap}. Mimicking human behaviors or gaits can simplify reward tuning, enhance social acceptance, and foster more natural human-machine interactions, paving the way for robots to act not just as tools but as intuitive companions.

Today, many approaches require procedurally complex multistage training \cite{zhang2025track, he2025asap}, and distilling multiple skills or policies into a single policy remains an open problem \cite{zhang2025track, taira2025hover}. Although RL has been applied to legged locomotion for decades \cite{russ, stone}, a substantial leap in robustness was first achieved on quadrupeds \cite{hwangbo2019learning, lee2020learning}, and since then, the same core algorithms have shown transferability to a wide range of morphologies \cite{bohlinger2024onepolicy}. However, the choice of morphology introduces a fundamental trade-off. Adding legs increases coordination demands but expands the space of recovery maneuvers. Quadrupedal robots reduce mechanical complexity introduced by complicated linkages and mechanisms required for a humanoid. Furthermore, bipedal locomotion relies on dynamic stability and is particularly challenging on rough terrain; however, it provides the benefit of a smaller footprint, and the humanoid form factor facilitates the scalable collection of human reference motion data.

In general, verifying and certifying the safety of control policies for legged robots remains an open problem. For this, popular approaches, reviewed in \cite{safe2022}, include Hamilton-Jacobi (HJ) reachability analysis, control barrier functions, and Lyapunov-based stability methods. Control barrier functions have been applied to bipedal navigation \cite{agrawal2017discrete}, and HJ reachability analysis has been used to reduce safety violations for learning-based quadrupedal navigation \cite{hsu2023sim}. Despite these advances, a substantial gap remains between the locomotion behaviors that can be achieved in practice and those for which formal safety guarantees can be established. In particular, defining safe states under partial observability and complex nonlinear environment dynamics remains difficult. These challenges continue to drive research in improving both safety and interpretability across different timescales, motivating ongoing research in hybrid systems, and providing statistical safety guarantees.

Because legged robot locomotion has advanced from smooth terrains to rough and then to discontinuous terrains \cite{shixin2024pie,zhuang2024humanoid, jenelten2024dtc}, the current focus is moving beyond pure geometric perception toward incorporating visual and semantic understanding \cite{escontrela2025gaussgym}. Humans themselves plan the next few footholds using their eyes, focusing their gaze on the most relevant areas of the terrain while interpreting various cues that influence how we locomote \cite{matthis2018gaze}. By using semantics, we can anticipate terrain interactions and deal with unstable terrain (loose stones and branches) and are mindful of the forces we exert on the environment. We term this emerging research paradigm, which emphasizes the need for robots to anticipate interactions with the terrain beyond mere geometry and interpret their environment to plan precise, fine-grained motor responses, ``dexterous semantic locomotion.'' This paradigm requires taking into account affordances that can be identified from multimodal data and used for dexterous motion control. Furthermore, developing locomotion behaviors that involve long-horizon planning and enable effective collaboration with humans and other agents is becoming increasingly important.

\begin{figure*}
	\centering
	\includegraphics[width=\textwidth]{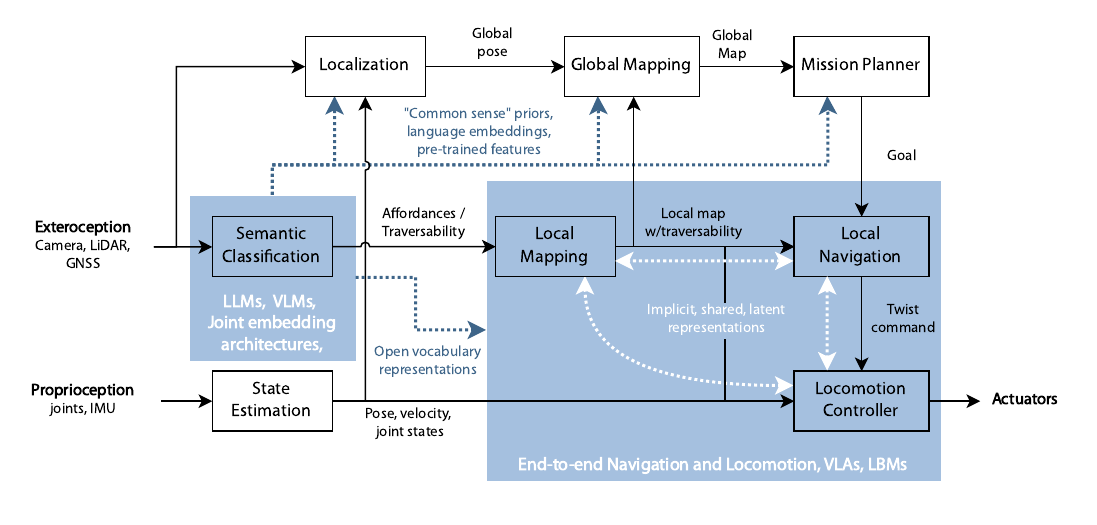}
	\caption{\footnotesize \textbf{Evolution of autonomy systems.} Classical autonomy systems decompose functionality into distinct submodules and process sensory information hierarchically, using clearly defined interfaces (illustrated as gray bounding boxes). Learning-based autonomy systems, by contrast, introduce tightly coupled or fused modules, illustrated in blue. Large language models (LLMs), visual-language models (VLMs), and other joint-embedding architectures enable holistic information extraction. This allows representations to become accessible throughout the entire architecture (dotted lines), replacing strictly hierarchical data flows. The integration of large pretraining, language priors, and reasoning capabilities allows moving beyond fixed and heuristically defined ontologies toward more flexible general-purpose autonomy. However, fundamental engineering principles, such as required control frequencies and sensor update rates, continue to shape the underlying system architecture. IMUs, inertial measurement units; VLA, visual-language-action models; LBM, large behavior models; GNSS, global navigation satellite system.}
	\label{fig:autonomy}
\end{figure*}

\section*{Autonomy and Representations}
\label{sec:autonomy}
Autonomy is a critical component of any robotic system. In the context of legged robots, an autonomy system generates autonomous navigation behavior, exploiting the legs and locomotion algorithms to negotiate obstacles and challenging terrain. Although legged robots are capable of more complex manipulation and interaction, here we focus primarily on navigation and refer the reader to \cite{kroemer2021review} for an exhaustive review of manipulation.

Traditionally, autonomy systems decompose the overall task into multiple subsystems and leverage hierarchical data processing with clearly defined interfaces to manage the complexity associated with autonomy (see Fig.~\ref{fig:autonomy}). The perception subsystem of a legged robot allows it to understand its own state and the environment. The resulting state estimates may include contact states, external forces, and ground reaction forces and can be used by locomotion control, navigation, mapping, and other modules. Estimation of the robot's pose and velocity, known as odometry, has been a major focus of development. Apart from mobility, legs serve as sensors, providing information about the robot's kinematics via leg odometry \cite{camurri2026legodometry}. Current-state estimators for legged robots rely on multisensor fusion, which has evolved from kinematic-inertial formulations \cite{bloesch2012state} to current solutions integrating cameras, LIDAR, radar, and other sensor modalities in complementary ways \cite{khattak2020compslam, jung2024coral, nubert2025holistic}. The resulting estimates are used to spatially and temporally fuse sensory information into structured representations that encode the geometry and semantics of the environment \cite{bradley2015Sceneunderstanding}, such as local terrain maps or global simultaneous localization and mapping maps, used for navigation.

Semantic understanding on a legged robot goes beyond merely determining the geometry and semantic labels of the environment. Instead, navigation systems must also consider the robotic hardware and controller in their analysis, a problem known as traversability estimation. Traversability estimation methods score regions in a terrain map or occupancy grid depending on how easy or hard it is for the robot to move across \cite{fan2021step}. The scores (or costs) are usually determined by a learned model, trained from manual labels or self-supervision signals \cite{wellhausen2019where, frey2023wvn}, which can be directly used by traditional motion planners to find feasible paths. However, the use of structured representations makes the integration of the robot's state and proprioceptive signals challenging. This challenge has motivated works that learn forward dynamics models to encode traversability without explicit scene representation \cite{roth2025fdm}.

Similarly to locomotion research, training navigation policies using RL has shown promising results. These methods learn directly from raw sensor inputs or from map representations \cite{lee2024autonomous, chong2024resilient, zhuang2023visionproprioception}. Nevertheless, replacing the navigation module with a learned version still keeps navigation and locomotion decoupled, which limits the deployment of legged robots in more challenging settings.

In contrast with simply replacing or enhancing individual components of the classical autonomy system with learned ones, emerging learning-based autonomy systems aim to merge different modules to overcome the limitations of hierarchical sensor information processing, hand-crafted interfaces, and heuristic-based representations (see highlighted blocks in Fig.~\ref{fig:autonomy}). This could lead to tighter integration of learned abstractions from vision and language, enabling fusion and mutual conditioning among modules through diverse explicit and implicitly learned representations.

A first step toward this alternative autonomy design is found in robot parkour, a use case motivated by rescue operations, where the traditional decoupling will likely fail. In this setting, the robot needs to navigate to a goal located in discontinuous and uneven terrain, facing gaps, high obstacles, and narrow spaces. Recent works have succeeded in coordinating different locomotion behaviors via hierarchical RL and latent scene representations \cite{hoeller2024anymalparkour, zhuang2023visionproprioception}.

Throughout the full autonomy system, we observe that abstractions learned from language play a key role, enabling reasoning over corner cases and high-level planning \cite{cheng2024navila, pmlr-v305-ginting25a}, capabilities that are essential for more complex and ``in-the-wild'' deployment. The learned semantic representations are maintained in diverse forms of memory, such as scene graphs \cite{pmlr-v305-ginting25a}, which allow retraining, grouping, and retrieval of abstract concepts and raw sensory information while still leveraging established navigation graphs approaches \cite{tranzatto2022cerberus, fan2021step}.

Complementary diffusion models \cite{Chi-RSS-23, reuss2023goal}, visual-language-action models \cite{kim2024openvla, gemini2025robotics}, and large behavior models \cite{lbmtri2025} have demonstrated generalization and dexterous manipulation capabilities. When pretrained on large-scale internet and robot teleoperation datasets, these models can learn effective behaviors from only a few demonstrations \cite{lbmtri2025}. These approaches offer a powerful paradigm for learning navigation behaviors. However, current navigation methods based on the imitation learning paradigm still lag behind classical approaches in terms of robustness and interpretability and struggle to effectively integrate proprioceptive information compared to RL policies trained in simulation. Ongoing developments in navigation foundation models have enabled new, intuitive ways to convey navigation instructions to robots using natural language \cite{cheng2024navila} and demonstrate the potential of fusing previously separate elements into a single architecture, pushing the boundaries toward generalist humanoid robots \cite{bjorck2025gr00t}. Despite the advantages of a unified approach, hard real-time requirements continue to drive architectural design, highlighting a need for more research into how different frameworks facilitate adaptive computing on the basis of task complexity. Defining the optimal interfaces, control frequencies, and functional boundaries between components, if any are needed, remains a subject of active debate.

Furthermore, it remains an open question which priors should be explicitly encoded in the autonomy system versus relying on end-to-end approaches that leverage scalable learning objectives and large amounts of training data. Even though a modular autonomy stack has established itself as a foundation for autonomy, relying on decoupled perception, mapping, planning, navigation, and locomotion modules might be insufficient to tackle more complex tasks. Modern applications require unprecedented situational awareness, comprehensive world knowledge, and tight embodied coordination between their low- and high-level behaviors. Their success will not only rely on the availability of training data and the underlying learning algorithm but also on the chosen problem formulation.

\begin{figure*}
	\centering
	\includegraphics[width=1.0\textwidth]{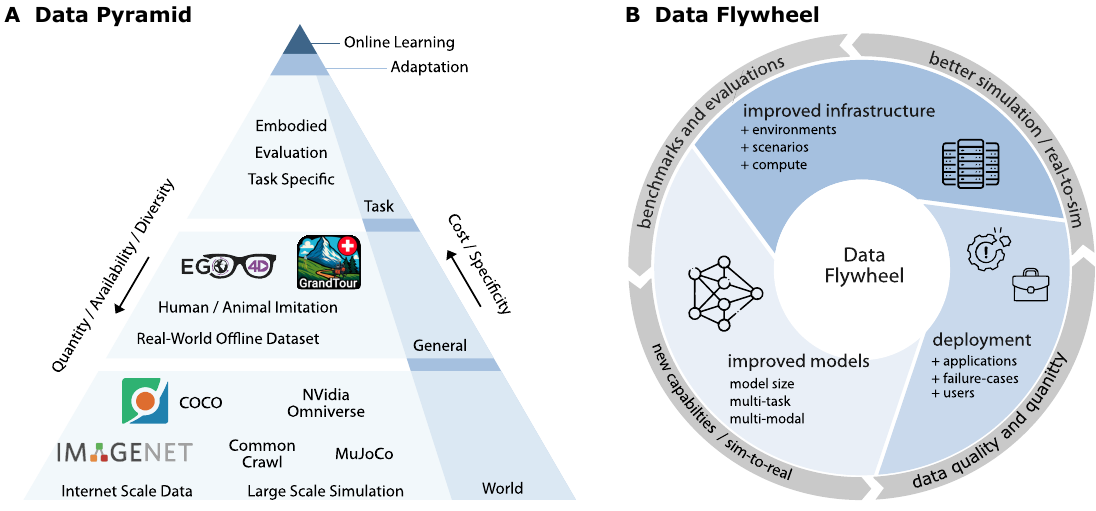}
	\caption{\footnotesize \textbf{Data and simulation overview.}
(\textbf{A}) The data scaling pyramid illustrates that data exist at varying levels of scale. The more task- and embodiment-specific the data are, the more costly they become to collect. Different data sources support learning different underlying concepts, which can be categorized as world knowledge, general knowledge, and task-specific knowledge.
(\textbf{B}) A modern data flywheel demonstrates that data collection should not occur in isolation but in a closed-loop system, where continuous feedback enables iterative improvement of model performance.
}
	\label{fig:simulation}
\end{figure*}

\section*{Data and Simulations}
\label{sec:data}
Acquiring suitable training and evaluation data is a fundamental challenge in developing legged systems designed to operate in diverse and unstructured environments. Data are often deeply tied to the robot's embodiment, sensor modalities, actuation dynamics, and environmental conditions. They are highly multimodal (in terms of sensor types), often asynchronous, and affected by time-correlated noise and system delays. These factors make large-scale, reliable data collection challenging, especially in the long tail of unpredictable edge cases where legged robots must operate robustly. In response, developers and researchers use diverse data sources that vary in specificity, quantity, and diversity to build different levels of abstraction and understanding. These range from world knowledge and general knowledge to task-specific understanding (compare the data pyramid illustrated in Fig.~\ref{fig:simulation}A).

The training of robust locomotion and navigation policies in simulation has been enabled by recent advances in graphics processing unit (GPU)--accelerated simulation frameworks \cite{makoviychuk2021isaacgym, zakka2025mujocoplayground, mittal2025isaaclab}, which democratized large-scale policy learning on consumer-grade hardware. These simulators allow for rapid data generation with privileged supervision signals, making them effective for learning control-oriented behaviors. However, simulation fidelity and realism often come at the cost of computational throughput. Approximations in physics modeling and visual rendering introduce a sim-to-real domain gap, limiting direct transfer of learned policies to hardware. Although perfect simulation remains elusive, techniques such as extensive domain randomization \cite{tobin2017domainrandomization}, learned residual models \cite{hwangbo2019learning, he2025asap}, and accurate system identification \cite{miller2025sysidspot} have proven effective in narrowing this gap. Consequently, most low-level locomotion, navigation, and whole-body control policies are now primarily trained in simulation, relying on proprioceptive signals and midlevel visual representations such as depth and semantic images.

Despite these advances, simulation alone is insufficient. Designing environments and modeling system uncertainties that expose the robot to a sufficiently diverse range of scenarios remains a manual and brittle process. Many situations are still beyond the reach of current simulators, for instance, entanglement of feet in vegetation, locomotion on deformable terrain, or the presence of semantically rich, cluttered environments. Moreover, generating high-fidelity RGB data is computationally expensive, typically requiring many GPUs \cite{escontrela2025gaussgym}. However, the main bottleneck is simulating the correct physical interaction corresponding to the visual data, which ultimately leads to the unresolved sim-to-real visual gap. These limitations highlight the importance of real-world datasets that capture the complexity, unpredictability, and semantic richness that simulation cannot yet replicate.

Real-world datasets for legged robots are obtained through direct teleoperation of the robot in different environments or indirectly via human or animal demonstrations. Motion references extracted from videos \cite{grauman2025ego4d} or motion capture systems \cite{mahmood2019amass, he2018modeadaptive} provide valuable priors for locomotion control, simplifying reward design and stabilizing imitation learning. For navigation and mapping, datasets with rich sensory payloads, including RGB-D cameras, LIDAR, and leg proprioceptive information, are essential for developing and benchmarking different algorithms. Unlike self-driving cars, collecting diverse legged robot data is much more expensive and requires robot deployment on complex, unstructured terrains rather than on roads. Recent efforts using quadrupedal robots, such as the GrandTour Dataset \cite{frey2025boxidesign} and the Sub-T dataset \cite{zhao2024subtmrsdataset}, have begun to close this gap by recording data in environments ranging from urban streets to forests and construction sites under varied lighting and weather conditions. Ensuring data integrity and verification throughout large deployments remains challenging, and the scale and diversity still fall short compared with those of autonomous driving. This makes real-world data a scarce yet crucial component complementary to simulation.

Bridging the gap between simulation and real-world data will play a central role in advancing legged robotics. Neural rendering techniques \cite{mildenhall2020nerf,kerbl2023gaussian,escontrela2025gaussgym} and neural scene reconstruction \cite{wang2025vggt} may soon enable large-scale, high-fidelity, photorealistic RGB rendering and meshing for simulation. Neural-augmented physics engines \cite{liu2021role} could better capture complex dynamics, extending coverage of long-tail failure modes that remain poorly modeled today. Improvements in differentiable simulation can support the development of more efficient learning algorithms, given that generative models \cite{ball2025genie, chen2025sam} provide a promising avenue for automatically constructing diverse, semantically rich environments, reducing the manual engineering burden. At the same time, broader deployment of legged robots, potentially in the beginning via human-guided teleoperation, will naturally expand real-world datasets (Fig.~\ref{fig:simulation}B). This can lead to internet-scale training paradigms similar to those already established in vision and language. Early evidence from robot manipulation underscores the benefits of larger-scale datasets \cite{lbmtri2025}. Combining scalable simulation, generative augmentation, and broad real-world deployment will be essential to create richly diverse datasets that can drive the next generation of robust, general-purpose legged systems.

Across all these modalities, a major bottleneck for researchers is the lack of standardized, shared benchmarks and practices. Most sim-to-real studies rely on customized hardware, making systematic comparison difficult \cite{caluwaerts2023barkour}. Simulated comparisons are also insufficient because they do not capture the challenges of transferring policies to the real hardware. These challenges require larger efforts beyond legged robotics, which could benefit the entire field. Ultimately, all of these advances can contribute to a data flywheel (Fig.~\ref{fig:simulation}B), which accelerates data availability, enhances simulation, improves benchmarking, and unlocks new real-world applications.

\begin{figure*}
	\centering
	\includegraphics[width=1.0\textwidth]{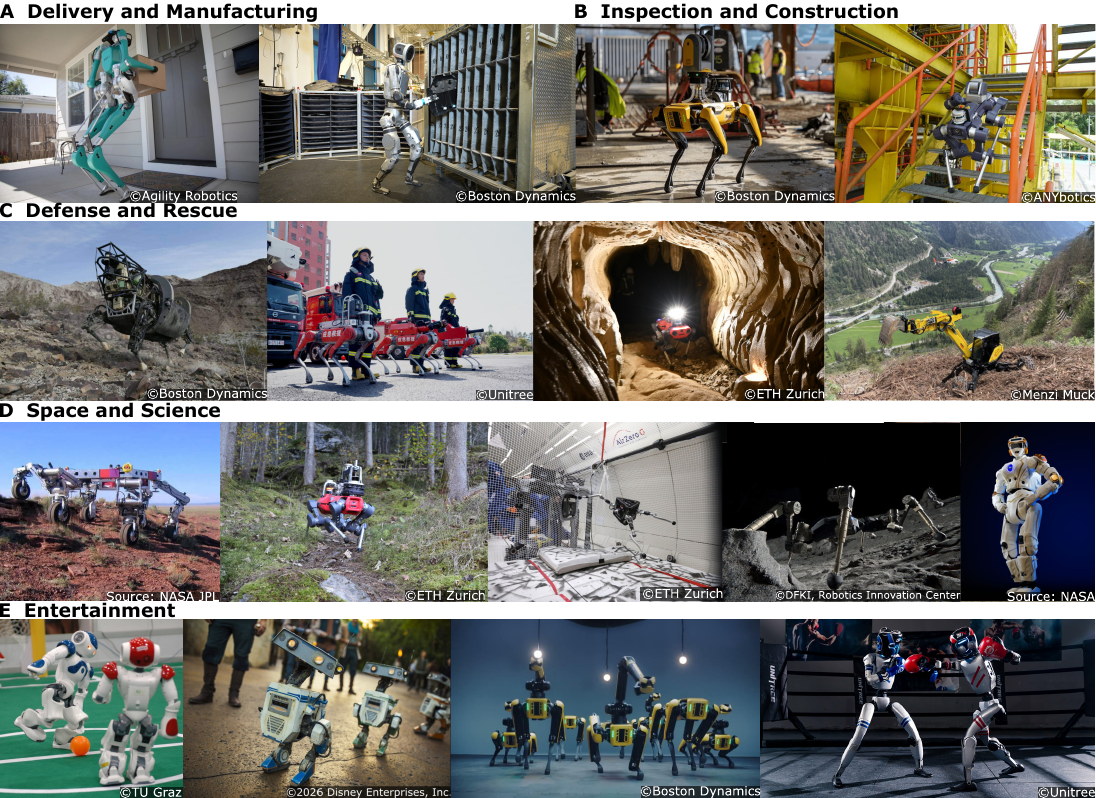}
	\caption{\footnotesize \textbf{Applications of legged robots.}
(\textbf{A}) Digit robot performs last-mile delivery, and Electric Atlas by Boston Dynamics performs mobile manipulation for an industrial task.
(\textbf{B}) Boston Dynamics' Spot monitors a construction site, and ANYmal D performs industrial inspection at a manufacturing plant.
(\textbf{C}) Boston Dynamics LS3 acting as a mule in rough terrain, Unitree B2 tested by a fire brigade for situational awareness and firefighting, ANYmal C exploring underground environments during the DARPA SubT challenge, and Menzi Muck walking excavator operating in a hard-to-access construction site.
(\textbf{D}) JPL ATHLETE rover for extraterrestrial exploration; ANYmal D in a forest monitoring application; SpaceHopper undergoing zero-gravity flight; SpaceClimber, one of the earliest legged robot prototypes for space exploration, developed at the German Research Center for Artificial Intelligence (DFKI); NASA Valkyrie humanoid robot operating as an astronaut.
(\textbf{E}) NAO robots playing soccer, Disney's believable robotic character, Boston Dynamics' Spot robots performing a synchronized dance routine, and Unitree's G1 boxing.
Image Credits: Agility Robotics, Boston Dynamics, ANYbotics Unitree, ETH Zurich, Menzi Muck, NASA, DFKI Robotics Innovation Center, TU Graz, Disney Enterprises, Inc..
}
	\label{fig:application}
\end{figure*}

\section*{Real-World Applications}
\label{sec:applications}
Historically, legged robots have been developed as an alternative to wheeled robots for transporting heavy loads on rough terrain \cite{todd1985walking}. However, the advances to date have widely expanded the possible use cases, affecting different economic sectors from production to entertainment (Fig.~\ref{fig:application}).

\subsection*{Economic sectors}
Inspection and monitoring has been the main business case for quadrupedal platforms. Robots such as the ANYbotics ANYmal have been deployed in offshore oil and gas facilities \cite{gehring2021ANYmal}, whereas the Boston Dynamics Spot has been adopted for inspection and monitoring of mines \cite{BostonDynamics_KiddCreekMine} and energy facilities, including nuclear \cite{staniaszek2025AutoInspect}. Long-term monitoring of construction sites for safety and progress assessment is another challenging application where these platforms are being deployed \cite{yue2025leggedrobots}.

Sectors that require operation in open fields, such as agriculture and forestry, introduce additional challenges for autonomy, namely, the lack of communication infrastructure, reference landmarks, and changing environmental conditions. Therefore, most of the demonstrations have been achieved in research projects. Some examples include agricultural monitoring \cite{quail2023agricultural}, forest harvesting \cite{jelavic2022roboticprecision}, and forest inventory \cite{mattamala2025building}. These deployments have focused on quadrupedal platforms, because they are inherently more stable than humanoids.

Delivery has recently become another commercial application for legged systems, targeting the ``last mile'' problem. Companies such as RIVR have demonstrated parcel and food delivery to real homes \cite{rivr2025}, showing the viability of combining teleoperation and shared autonomy for van-to-door deliveries in diverse urban settings.

The new ``humanoid boom'' started in 2022 and aims to deploy these platforms in factories, supporting manufacturing and load-carrying tasks in warehouses. Even though this was considered an ``unsuitable'' task for legged platforms in the past \cite{todd1985walking}, the private sector has shown rapid progress in building general-purpose robots that can potentially take over a large variety of tasks using the same hardware and software. Commercial pilot humanoids promise to perform simple tasks in private domestic settings by 2026 \cite{Jang2025_NEOHomeRobot}.

Another long-standing use case for humanoid robots is supporting human carers in activities of daily living. Their morphology makes them well suited to integrate into home environments and allow for intuitive interaction. Japan has led developments in this direction, motivated by its aging population \cite{miyake2025dualarmmotion}. Nevertheless, the main challenges in this domain are not related to mobility but instead to compliant manipulation, safe human interaction, and scene and intention understanding.

\subsection*{Defense and disaster response}
The need for ``robotic mules'' to carry heavy loads during military field operations was the key motivation for developing the Boston Dynamics BigDog and LS3 in the late 2000s. This period involved the largest field deployments ever reported for a legged platform \cite{bradley2015Sceneunderstanding}.

Using humanoid robots to test chemical-resistant military gear led to the development of the Atlas humanoid \cite{bostondynamics2016atlas}, which was later used in the DARPA Robotics Challenge (DRC). The DRC focused on solving tasks relevant to disaster response, including rough terrain traversals, tool manipulation, and vehicle driving. The challenge inspired follow-up research projects in areas such as firefighting \cite{tsagarakis2017walkman}, but it also highlighted hardware and control limitations in humanoid robots of that era. These shortcomings motivated a shift in the late 2010s toward more dynamically stable platforms, such as quadrupeds.

The DARPA Subterranean Challenge (SubT), held from 2019 to 2021, focused on underground exploration using heterogeneous robotic teams. This became a key demonstration of the maturity of quadrupedal robots, by being part of most of the finalist teams \cite{chung2023darpa}. A central challenge in both subterranean and combat environments is the high demand for autonomous decision-making, because radio signals can be jammed or high-bandwidth communication cannot be established underground. Wired communication can provide the required bandwidth, but it is limited in range and vulnerable to adversarial attacks. Therefore, legged robots require advanced autonomy without human-in-the-loop control. Today, military applications represent one of the clearest paths to commercialization because of the strong demand for advanced mobility, but this also raises important ethical questions (see Ethics). Still, SubT has helped kick off several civil research and commercial applications in forestry, agriculture, and delivery, as discussed previously.

\subsection*{Scientific applications}
Quadruped and biped robots have been historically used as physical models to understand locomotion \cite{shin2024fastgroundtoair}. Now, their advanced mobility skills have motivated their use as data acquisition tools for environmental sciences, such as soil sampling \cite{wilson2021spatiallyandtemporally}, and as monitoring tools for animal interaction in the wild \cite{melo2023africanwilderness, canteloup2024monkeys}.

Space and planetary sciences have formed key use cases for legged robots since the 1960s \cite{todd1985walking}. NASA's LEMUR project studied the applicability of using legged robotic platforms in unstructured environments, because legs enable walking, crawling, and climbing \cite{parness2017lemur3}. More recently, the European Space Agency Space Resources Challenge showed how quadruped platforms propose alternative strategies for scientific exploration where wheeled rovers might be inappropriate, such as caves or craters \cite{arm2023scientificexploration}. However, critical issues, such as thermal and energy management, as well as reliability, remain major challenges.

\subsection*{Art and entertainment}
The agility of legged robots is an appealing characteristic for arts and entertainment. Consumer platforms such as the AIBO (Artificial Intelligence RoBOt), QRIO (Quest for cuRIOsity), and NAO have been widely used for social interaction and art performances. The RoboCup \cite{RoboCupSPL2025} is one of the biggest efforts that combines a research challenge and an annual world competition, motivating recent initiatives such as the World Humanoid Robot Games 2025 \cite{WorldHumanoidRobotGames2025}.

The increased hardware and control robustness of modern legged platforms has recently sparked their use as embodiments of popular movie characters. Disney's bipedal robot is an exemplar demonstration, being successfully deployed in parks and live shows \cite{grandia2024design}. This also opens opportunities for exploring expressive motion behaviors beyond locomotion or manipulation, which can be critical for human interaction and engagement.

\section*{Ethics}
\label{sec:ethics}
For the most part, legged robots present ethical issues that are not fundamentally different from those associated with other types of robots \cite{lin2014robot}. However, some of these arise with greater force for legged robots owing to the applications in which legged robots are likely to be used.

The capacity of robots to perform tasks that have traditionally been carried out by human beings generates a risk of technological unemployment. This is to be regretted because (some) paid work also provides people with an opportunity to contribute meaningfully to their community. Loss of jobs to robots is also likely to widen social and economic inequality and may provoke social unrest. These latter possibilities have driven interest in redistributing some of the wealth generated by robots via a ``universal basic income'' \cite{ford2015rise}.

The possibility that developments in robotics may lead to large social and economic changes highlights that there are ethical and political questions about how, and by whom, decisions about the development of robots are made. If a politician announced that they were going to radically change society, the public would demand a say in the matter. The fact that engineers, and those who fund them, are overwhelmingly men, from a relatively narrow range of sociodemographic backgrounds, also foregrounds questions about the extent of the democratic mandate for their projects \cite{sparrow2023robotspolitics}.

The development of humanoid robots brings us closer to the longstanding dream of a robot ``butler.'' Research to develop robot butlers is often justified by pointing to a looming ``demographic crisis''---that means that there will be fewer human carers available to look after a larger population of older persons. It is unclear whether older people want to be looked after by robots and whether ``care'' provided by robots can meet the social and emotional needs of older persons \cite{sparrow2006inthehands}. Ethical questions related to access to, and control over, the data gathered by sensors on robots arise with special force in relation to robots designed for applications in the home and in health- and aged-care settings.

Providing robots with legs may expand the number of roles that robots can take on in war \cite{scharre2018armyofnone}. Critics have alleged that military robots lower the psychological barriers to killing and lower the threshold of conflict and that the use of ``autonomous weapon systems'' would problematize the allocation of responsibility for their effects and violate the human rights of enemy combatants \cite{sparrow2016robotsandrespect}.

Until robots become sentient---if they ever do---the ethics of the way people treat robots will be determined by what it reveals about us and/or how it affects other people. Robots that look like people or animals, as may legged robots, have a greater capacity to encourage users to form emotional bonds with them that may generate benefits for, or lead to harms to, the users. Whether real friendship is possible with things that have no feelings is contested, as is the extent to which ersatz relationships contribute to well-being \cite{sparrow2016robotsinagedcare}. Nevertheless, it seems likely that people might experience real harm if their relationships with robots end unexpectedly when, for instance, a robot breaks down or is destroyed. Conversely, it is possible that by encouraging people to replace relationships with people with relationships with robots, the development of companion robots may increase social isolation \cite{sparrow2026against}. There is also a danger that the lessons users learn from their interactions with robots may affect the way they relate to people and/or animals \cite{coghlan2019couldsocialrobots}.

The way we treat robots that look like people or animals may also say more about us than the way we treat other robots. Kicking a robot dog may reveal us to be cruel in a way that kicking a robot vacuum cleaner does not \cite{sparrow2021virtueandvice}. A user's mistreatment of ``female'' robots may send a message about their attitudes toward real women \cite{sparrow2017robots}. More generally, legged robots will raise issues of what might be called ``robot media ethics,'' especially in relation to how they represent people \cite{sparrow2023dorobotshavesex}. In particular, designers of humanoid robots should be aware of the complex race politics of robots \cite{williams2025degrees}. Historically, robots were first conceived of as ``mechanical slaves,'' and so, racially coded as Black, but contemporary images of humanoid robots, if not always the robots themselves, almost always have white, or gleaming metallic, surfaces, suggesting that they are now racially coded as white \cite{sparrow2020dorobotshaverace}. The gender politics of robots is also complex, given the role played by the fantasy of creating an ``artificial wife'' in the history of robotics \cite{wosk201myfairladies}. The more robots walk among us, the more important these, and other, ethical questions will become.

\section*{Policy and Economics}
\label{sec:policy}
The emergence of advanced legged robots further presents implications for labor markets and regulatory frameworks. Understanding the economic trajectory of legged robots through capability progression, from current walking-only systems to future manipulation and social interaction, permits us to map out distinct market phases and policy challenges.

\subsection*{Economic repercussions}
Unlike other robots, legged robots enable operation in both rough terrain and in environments designed for human legs, making such robots a compelling candidate for the automation of work. Current commercially available walking-only quadrupedal robots like Unitree B2 and Boston Dynamics Spot, available at a price range of \$30,000 to \$90,000 (as of 2025), serve niche markets in delivery, inspection, entertainment, and security, constrained by battery life (90 min to 6 hours) that limits range to 4 to 20 km. Smaller-scale quadrupeds and humanoids with entry-level models starting at \$2700 and \$4900 (as of 2025) suggest rapid commoditization \cite{unitree2020a1, unitree2024r1}. The rapidly developing capability progression will unlock escalating market potential from walking-only, to manipulation in manufacturing; manipulation for household tasks; and social capabilities enabling entry into large markets like elderly care.

It has been estimated that 47\% of US employment faces high automation risk from emerging technologies, including robots \cite{frey2017futureofemployment}, and it has also been found that one additional industrial robot per 1000 workers reduces wages by 0.42\% \cite{acemoglu2020robotsandjobs}. However, legged robots distinctly differ from industrial robots. Legged robots, initially, seem likely to create new markets such as household robotics \cite{Jang2025_NEOHomeRobot} and robotic last-mile delivery \cite{rivr2025}, as a result of both their novelty and distinct capability limitations. However, as legged platforms develop more capable manipulation and social interaction abilities, these systems may serve as direct substitutes for human labor. Importantly, unlike industrial robots, which affect only the $\sim$7.5\% of workers in the manufacturing sector, legged robots have the potential to affect service sectors, which collectively use $\sim$80\% of some developed economies' workforces \cite{BLS_2024}.

\subsection*{Global competition and regulatory divergence}
Current global regulatory approaches reveal competing philosophies. The European Union's (EU's) AI Act \cite{EU_AI_Act_2024}, effective August 2024, classifies robotics by risk level, requiring extensive compliance for high-risk applications. Japan's Society 5.0 \cite{Japan_Society5_2016} instead takes a promotional stance, investing to create a ``super-smart society'' where robots assist in elderly care and manufacturing. China's approach combines aggressive industrial policy with explicit targets, given that the Ministry of Industry and Information Technology aims for mass production of humanoid robots by 2025 and world leadership by 2027 \cite{China_MIIT_2023}. The United States lacks comprehensive legislation, relying on sectoral guidelines.

Such regulatory divergence is likely to matter economically. The International Federation of Robotics \cite{IFR_2024} reports that 4.28 million industrial robots operate globally, with China installing 276,288 units (51\% of global total) and Asia accounting for 70\% of deployments. China's industrial robot density has reached 470 per 10,000 employees, surpassing Germany and Japan, and a substantive price differential exists between Asian and Western robotics platforms \cite{IFR2024RobotDensity}. These facts suggest that coordinated national strategies and a domestic supply chain can provide competitive advantages.

\subsection*{Future outlook}
The next decade demands proactive policy responses as capabilities advance from mobility to manipulation to social interaction. We highlight four policy priorities.

First, capability-based regulation should scale requirements with robot abilities---likely minimal oversight for walking-only systems, moderate for domain-specific manipulation, comprehensive for general manipulation and social interaction. Such an approach prevents the stifling of innovation while ensuring protection as risks increase. The EU's risk-based approach provides a foundation, although it may need calibration to avoid disadvantaging Western manufacturers against Asian competitors.

Second, international coordination is a priority when emerging capabilities introduce safety risks. Special mention must be made of the potential for lethal legged robots, with the autonomous weapons framework in \cite{UN_LAWS_2024} providing governance models. More broadly, international standards such as \cite{ISO_TC299_2024} must continue to expand in response to future risks.

Third, strategic industrial policy must address competitive dynamics: Western nations risk technology dependency without intervention. Rather than protectionism, support should focus on research and development funding, manufacturing infrastructure, and skills development.

Last, anticipatory workforce programs must recognize compressed adjustment time frames. Unlike previous automation waves, which spanned generations, the plausible 10- to 15-year progression from robot walking to robot social capabilities demands rapid adaptation. Programs should identify and foster complementary human skills at each capability level: fleet management and technical support today, robot team-building tomorrow.

The economic potential of legged robots hinges on policy choices. Nations must aim to combine strategic investment, adaptive regulation, and workforce development. The convergence of falling hardware costs, increasingly powerful artificial intelligence (AI), and expanding applications suggest that legged robots represent transformative technology---governance innovation must aim to match these technical innovations.

\section*{Conclusion}
\label{sec:conclusion}
This Review presents the past, current, and potential future of legged robots. Driven by electromagnetic actuation, legged hardware has matured to the point of broad applicability. Quadrupedal robots are already commercially available from multiple companies, showing convergence in morphology and design. Humanoid hardware is evolving rapidly amid surging academic, governmental, and industrial investment; multiple competing designs coexist, and mass production is ongoing or planned for 2026 across many companies.

RL has made quadrupedal locomotion an approachable problem, yielding reliable performance across diverse environments, especially when terrain geometry is known. The frontier has now shifted toward semantic understanding and dexterous interaction. For bipedal robots, purely reactive behaviors are insufficient: Many tasks require precise foot placement and nuanced environment understanding. Meeting these challenges may require rethinking the sim-to-real paradigm, moving beyond rigid-body simulation, and developing new learning-based control algorithms. Because these dexterity issues mirror those in manipulation, they open opportunities for cross-domain advances.

Autonomy for legged robots, which is fundamental for unlocking large economic and social value, will be driven by learning and data, overcoming teleoperation and brittle heuristic approaches. Therefore, a key unsolved challenge toward autonomy and general-purpose robots is learning from diverse data sources and integrating the right inductive biases to handle the complexity of the real world, which will shape the future architecture of deployed autonomy stacks. Furthermore, the ongoing investments in legged robots, coupled with their general-purpose nature, position them uniquely to bridge the gap between digital AI agents and the physical world, providing a foundation for grounding intelligence in real-world interaction.

Even though many of today's industrial applications for humanoid robots on flat factory floors do not require advanced mobility, these deployments can represent an important first step toward general-purpose robots. When combined with economies of scale and mass production, such robots may become a viable solution even for applications that initially seem counterintuitive. To introduce this technology safely and responsibly into society, we advocate for capability-based governmental regulation, international coordination, strategic industrial policy, and anticipatory workforce programs.

At the same time, ethical considerations must be reflected in legislation, and we urge developers and researchers to remain vigilant about the implicit biases and potential harms that legged robots can introduce. Consequently, we must carefully consider the critical questions of who holds the authority to develop, deploy, and control these technologies.

\section*{Acknowledgments}
\label{sec:ack}
We thank V.~Ferrari for valuable help in editing this work.
Generative AI technology was used in the preparation of this document to improve grammar and refine text. All authors retain full responsibility for the content and its presentation.

\section*{Funding}
\label{sec:funding}
M.~Mattamala was funded by the UKRI Turing AI World Leading Researcher Fellowship on AI for Person-Centred and Teachable Autonomy (grant EP/Z534833/1).
M.~Mittal was funded by NVIDIA and the Swiss National Science Foundation through the National Centre of Competence in Automation (NCCR automation).

\section*{Author contributions}
\label{sec:contributions}
J.F. led the project; organized the survey; wrote the Introduction, Locomotion, Autonomy and Representations, and Conclusions sections; and prepared the figures.
M.~Mattamala organized the survey and wrote the Introduction, Autonomy and Representations, and Real-World Applications sections.
J.F. and M.~Mattamala extensively revised the manuscript and integrated the different contributions of each author.
H-W.P. wrote the Hardware section and contributed to the associated figures.
M.~Mittal wrote the Data and Simulation section and contributed to the associated figure.
M.O. wrote the Policy and Economics section.
R.S. wrote the Ethics section.
G.M. and M.H. provided guidance, input to selected sections, and comments on revisions.
All authors revised and approved the manuscript.

\bibliography{references}
\bibliographystyle{sciencemag}

\end{document}